\documentclass[runningheads]{llncs}
\usepackage{graphicx}

\usepackage{tikz}
\usepackage{comment}
\usepackage{amsmath,amssymb} 
\usepackage{color}
\usepackage{booktabs}       
\usepackage{multirow}
\usepackage{subfigure}
\usepackage[accsupp]{axessibility}  

\graphicspath{ {images/} }

\makeatletter 
\newcommand\figcaption{\def\@captype{figure}\caption} 
\newcommand\tabcaption{\def\@captype{table}\caption} 
\makeatother

\usepackage{listings}
\usepackage{color}

\definecolor{dkgreen}{rgb}{0,0.6,0}
\definecolor{gray}{rgb}{0.5,0.5,0.5}
\definecolor{mauve}{rgb}{0.58,0,0.82}

\begin{document}
	\pagestyle{headings}
	\mainmatter
	\def\ECCVSubNumber{334}  
	
	\title{Spatiotemporal Self-attention Modeling with Temporal Patch Shift for Action Recognition} 

	\titlerunning{TPS}
	%
	\author{Wangmeng Xiang\inst{1}$^{,}$\inst{2}\thanks{Work done during an internship at Alibaba.} \and
		Chao Li\inst{2} \and
		Biao Wang \inst{2} \and
		Xihan Wei\inst{2} \and
		Xian-Sheng Hua\inst{2} \and
		Lei Zhang\inst{1}\thanks{Corresponding author.}
	}
	\authorrunning{W. Xiang et al.}
	%
	\institute{The Hong Kong Polytechnic University, Hong Kong SAR, China \\
		\email{\{cswxiang,cslzhang\}@comp.polyu.edu.hk}\\ \and
		DAMO Academy, Alibaba, Hangzhou, China\\
		\email{\{lllcho.lc,wb.wangbiao,xihan.wxh,xiansheng.hxs\}@alibaba-inc.com}}
	\maketitle
	
	\begin{abstract}
		
		Transformer-based methods have recently achieved great advancement on 2D image-based vision tasks. For 3D video-based tasks such as action recognition, however, directly applying spatiotemporal transformers on video data will bring heavy computation and memory burdens due to the largely increased number of patches and the quadratic complexity of self-attention computation. How to efficiently and effectively model the 3D self-attention of video data has been a great challenge for transformers. In this paper, we propose a Temporal Patch Shift (TPS) method for efficient 3D self-attention modeling in transformers for video-based action recognition. TPS shifts part of patches with a specific mosaic pattern in the temporal dimension, thus converting a vanilla spatial self-attention operation to a spatiotemporal one with little additional cost. As a result, we can compute 3D self-attention using nearly the same computation and memory cost as 2D self-attention. TPS is a plug-and-play module and can be inserted into existing 2D transformer models to enhance spatiotemporal feature learning. The proposed method achieves competitive performance with state-of-the-arts on Something-something V1 \& V2, Diving-48, and Kinetics400 while being much more efficient on computation and memory cost. The source code of TPS can be found at https://github.com/MartinXM/TPS.
		
		\keywords{action recognition, transformer, temporal patch shift}
	\end{abstract}

	\section{Introduction}
	
	Significant progresses have been achieved for video based action recognition in recent years~\cite{carreira2017quo,qiu2017learning,feichtenhofer2019slowfast,wang2016temporal,lin2019tsm}, largely driven by the development of 3D Convolutional Neural Networks (3D-CNN) and their factorized versions, including I3D~\cite{carreira2017quo}, Slowfast~\cite{feichtenhofer2019slowfast}, P3D~\cite{qiu2017learning}, TSM~\cite{lin2019tsm}. With the recent success of transformer-based methods on image based tasks such as image classification, segmentation and detection~\cite{dosovitskiy2021an,liu2021swin,yuan2021tokens,han2021transformer,touvron2021training,chen2021crossvit}, researchers have been trying to duplicate the success of transformers on image based tasks to video based tasks~\cite{GedasBertasius2021IsSA,AnuragArnab2021ViViTAV,ZeLiu2021VideoST}. Specifically, videos are tokenized as 3D patches, then multi-head Self-Attention (SA) and Feed-Forward Networks (FFN) are utilized for spatiotemporal feature learning. However, the extra temporal dimension of video data largely increases the number of patches, which leads to an exponential explosion in computation and memory cost as the calculation of multi-head SA has a quadratic complexity.

	\begin{figure}[t]
		
		\begin{center}
			\includegraphics[width=0.8\textwidth]{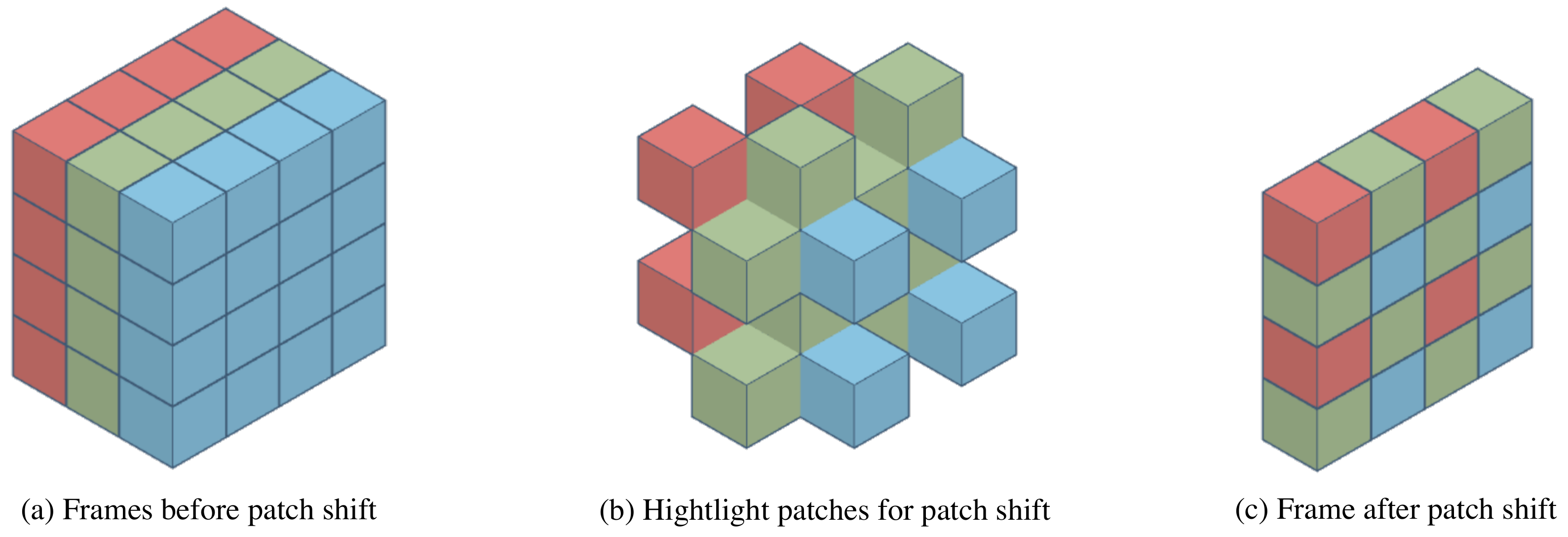}		
			\includegraphics[width=0.9\textwidth]{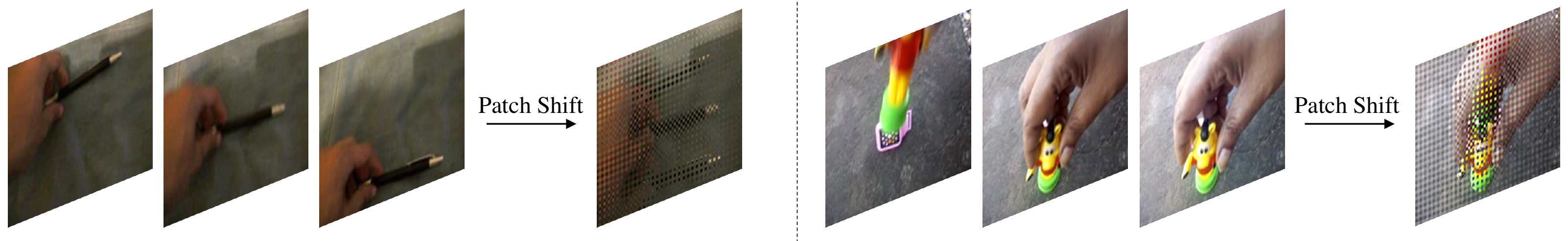}
			\caption{An example of temporal patch shift for three adjacent frames.}
			\label{fig:frontpage}
		\end{center}
	\end{figure}
	
	Previous efforts to reduce the computational burden of spatiotemporal multi-head SA are mainly focused on how to factorize it into spatial and temporal domains and compute them separately~\cite{GedasBertasius2021IsSA,AnuragArnab2021ViViTAV}. For example, Timesformer~\cite{GedasBertasius2021IsSA} first applies spatial-only SA and then temporal-only SA in a transformer encoder. ViViT~\cite{AnuragArnab2021ViViTAV} adds a few temporal-only transformer encoders after spatial-only encoders. However, these factorization methods will introduce additional parameters and computation for temporal SA calculation comparing to a spatial-only transformer network.

	With the above discussions, one interesting question is: Can we endow 2D transformers the capability of temporal SA modeling without additional parameters and computational cost? To answer this question, we propose a Temporal Patch Shift (TPS) method for efficient spatiotemporal SA feature learning. In TPS, specific mosaic patterns are designed for patch shifting along the temporal dimension. The TPS operation is placed before the SA layer. For each frame, part of its patches are replaced by patches from neighboring frames. Therefore, the current frame could contain information from patches in temporal domain, and the vanilla spatial SA module can be extended to a spatiotemporal one. It is worth noting that, although the spatiotemporal self-attention computed by TPS is sparse, the spatiotemporal receptive field can be naturally expanded as the TPS layers are stacked. A special case of TPS, where the patches from neighboring frames are shifted using a ``Bayer filter", is shown in the first row of Fig. \ref{fig:frontpage}. We highlight the patches that are being shifted. It can be seen that, by replacing half of the patches in the current frame with patches from previous and next frames, the vanilla spatial SA is upgraded to spatiotemporal SA with a temporal receptive filed of 3. In the second row of Fig.~\ref{fig:frontpage}, we show two examples of visualization for consecutive frames after patch shift, which indicate the motion of actions can be well presented within a single frame. The contributions of this work are summarized as follows:
	\begin{itemize}
		\item We propose a Temporal Patch Shift (TPS) operator for efficient spatiotemporal SA modeling. TPS is a plug-and-play module and can be easily embedded into many existing 2D transformers without additional parameters and computation costs. 
		\item We present a Patch Shift Transformer (PST) for action recognition by placing TPS before the multi-head SA layer of transformers. The resulted PST is highly cost-effective in both computation and memory.
		
	\end{itemize}

	Extensive experiments on action recognition datasets show that TPS achieves $58.3\%$, $69.8\%$, $82.5\%$ and $86.0\%$ top-1 accuracy on Something-something V1 $\&$ V2, Kinetics400 and Diving48, which are comparable to or better than the best Transformer models but with less computation and memory cost.

	\section{Related works}
	
	Action recognition is a challenging and cornerstone problem in vision. Many deep learning based methods have been proposed recently~\cite{LSF3Diccv2015,CTNFcvpr2016,TCNnips2014,rohrbach2016recognizing,LSF3Diccv2015,yang2020temporal,piergiovanni2019representation,christoph2016spatiotemporal,martinez2019action,zheng2020dynamic,zhao2018recognize,li2020tea,feichtenhofer2019slowfast,wang2016temporal,TSN2016ECCV,carreira2017quo,qiu2017learning,lin2019tsm}. Based on the employed network architecture, they can be categorized into CNN-based ones and transformer-based ones. 
	
	\textbf{CNN-based methods.} CNN based methods typically use 3D convolution~\cite{LSF3Diccv2015,carreira2017quo,feichtenhofer2019slowfast} or 2D-CNN with temporal modeling~\cite{TSN2016ECCV,qiu2017learning,lin2019tsm} to construct effective backbones for action recognition. For example, C3D~\cite{LSF3Diccv2015} trains a VGG model with 3D-CNN to learn spatiotemporal features from a video sequence. I3D~\cite{carreira2017quo} inflates all the 2D convolution filters of an Inception V1 model~\cite{inceptionv1} into 3D convolutions so that ImageNet pre-trained weights can be exploited for initialization. Slowfast~\cite{feichtenhofer2019slowfast} employs a two-stream 3D-CNN model to process frames at different sampling rates and resolutions. Due to the heavy computational burden of 3D-CNN, many works attempt to enhance 2D-CNN with temporal modules~\cite{TSN2016ECCV,qiu2017learning,lin2019tsm,liu2020teinet,li2020tea,wang2021tdn}. P3D~\cite{qiu2017learning} factorizes 3D convolution to 1D temporal convolution and 2D spatial convolution. TSM~\cite{lin2019tsm} presents an efficient shift module, which utilizes \textit{left} and \textit{right} shifts of sub-channels to substitute a group-wise weight-fixed 1D temporal convolution. TEA~\cite{li2020tea} employs motion excitation and multiple temporal aggregation to capture motion information and increase temporal receptive field. TEINet~\cite{liu2020teinet} uses motion enhanced module and depth-wise 1D convolution for efficient temporal modeling. However, CNN-based methods cannot effectively model long-range dependencies within or cross the frames, which limits their performances.

	\textbf{Transformer-based methods.} Recently, with the advancement of transformers in 2D vision tasks~\cite{dosovitskiy2021an,liu2021swin,yuan2021tokens,han2021transformer,touvron2021training,chen2021crossvit}, many works have been done to apply transformers on video action recognition~\cite{GedasBertasius2021IsSA,AnuragArnab2021ViViTAV,ZeLiu2021VideoST,HaoZhang2021TokenST}. Different from the temporal modeling for CNN, which is mainly implemented by 3D convolution or its factorized versions, in transformers the spatiotemporal SA is naturally introduced to explore the spatiotemporal video correlations. Intuitively, one can use all the spatiotemporal patches to directly compute the SA. However, this operation is very computation and memory expensive. Many works are then proposed to reduce the computation burdens of joint spatiotemporal SA modeling. Timesformer~\cite{GedasBertasius2021IsSA} adopts divided space-time SA, which adds temporal SA after each spatial SA. ViViT~\cite{AnuragArnab2021ViViTAV} increases the temporal modeling capability by adding several temporal transformer encoders on the top of spatial encoders. Video swin transformer~\cite{ZeLiu2021VideoST} reduces both spatial and temporal dimension by using spatiotemporal local windows. Inspired by the temporal modeling methods in CNN, TokenShift~\cite{HaoZhang2021TokenST} enhances ViT for temporal modeling by applying partial channel shifting on class tokens.
	
	The success of temporal modules in 2D-CNN~\cite{qiu2017learning,lin2019tsm,liu2020teinet,li2020tea,wang2021tdn} motivates us to develop TPS to enhance a spatial transformer with spatiotemporal feature learning capability. Our work shares the spirits with TokenShift~\cite{HaoZhang2021TokenST} in terms of enhancing the temporal modeling ability of transformers without extra parameters and computation cost. However, our TPS is essentially different from TokenShift. TPS models spatiotemporal SA, while TokenShift is a direct application of TSM in transformer framework, which is in the nature spatial SA with ``temporal mixed token''. In addition, TPS does not rely on class token (not exists in many recent transformer models~\cite{liu2021swin}) and operates directly on patches, which makes it applicable to most recent transformer models.

	\section{Methodology}
	
	In this section, we present in detail the proposed Temporal Patch Shift (TPS) method, which aims to turn a spatial-only transformer model into a model with spatiotemporal modeling capability. TPS is a \textit{plug-and-play} operation that can be inserted into transformer with no extra parameters and little computation cost. In the following, we first describe how to build a visual transformer for videos, and then introduce the design of TPS for action recognition. 
	\subsection{Video-based Vision Transformer} 	
	The video based transformer can be built by extending the image based ViT~\cite{dosovitskiy2021an}. A video clip $X \in \mathbb{R}^{F\times H \times W \times C}$ can be divided into $s\times k \times k$ non-overlapped patches. The 3D patches are flattened into vectors $\mathbf{x}^{(t,p)} \in \mathbb{R}^{3sk^2}$ with $t = 1, \dots, T$ denoting the temporal index with $T = F/s$, and $p = 1, \dots, N$ denoting the spatial index with $N = HW/k^{2}$. The video patches are then mapped to visual tokens with a linear embedding layer
	
	\begin{equation}
		\begin{aligned}
			\mathbf{z}^{(t,p)}_{0}&=E\mathbf{x}^{(t,p)}+ \mathbf{e}_{pos}^{(t,p)}
		\end{aligned}
		\label{eq:embedding}
	\end{equation}
	
	\noindent where $E \in \mathbb{R}^{D\times 3sk^2}$ is the weight of a linear layer, $\mathbf{e}^{(t,p)}_{pos}$ is learnable spatiotemporal positional embedding, $\mathbf{z}^{(t,p)}_{0}$ represents an input spatiotemporal patch at location $(t,p)$ for transformer. We represent the whole input sequence as $\mathbf{Z}_0$.
	
	Suppose that a visual transformer contains $L$ encoders, each consisting of multi-head SA, Layer-Norm (LN) and FFN. The transformer encoder could be represented as follows:

	\begin{equation}
		\begin{aligned}
			\mathbf{\hat{Z}}_{l}&=\text{SA}(\text{LN}(\mathbf{Z}_{l-1}))+ \mathbf{Z}_{l-1}, \\
			\mathbf{Z}_{l}&=\text{FFN}(\text{LN}(\mathbf{\hat{Z}}_{l}))+ \mathbf{\hat{Z}}_{l}, \\
		\end{aligned}
		\label{eq:swin}
	\end{equation}
	
	\noindent where $\mathbf{\hat Z}_{l}$ and $\mathbf{Z}_{l}$ denote the output features of the SA module and the FFN module for block $l$, respectively. The multi-head SA is computed as follows (LN is neglected for  convenience):
	
	\begin{equation}
		\begin{aligned}
			Q_{l}, K_{l}, V_{l} & = W^{Q}_{l}\mathbf{Z}_{l-1},  W^{K}_{l}\mathbf{Z}_{l-1},  W^{V}_{l}\mathbf{Z}_{l-1}    \\
			\mathbf{\hat{Z}}_l &	= \text{SoftMax}(Q_{l}K_{l}^{T}/\sqrt{d})V_{l}, 
			\label{eq:SA}
		\end{aligned}
	\end{equation}
	
	\noindent where $Q_{l}$,$K_{l}$,$V_{l}$ represent the \textit{query}, \textit{key} and \textit{value} matrices for block $l$ and $W_{l}^{Q},W_{l}^{K},W_{l}^{V}$ are weights for linear mapping, respectively. $d$ is the scaling factor that equals to \textit{query}/\textit{key} dimension.
	
	Following transformer encoders, temporal and spatial averaging (or temporal averaging only if using class tokens) can be performed to obtain a single feature, which is then fed into a linear classifier. The major computation burden of transformers comes from the SA computation. Note that when full spatiotemporal SA is applied, the complexity of attention operation is $\mathcal{O}(N^2T^2)$, while spatial-only attention costs $\mathcal{O}(N^2T)$ in total. Next, we show how to turn a spatial-only SA operator to a spatiotemporal one with TPS.
	
	\subsection{Temporal Patch Shift} 
	
	\noindent \textbf{Generic shift operation.} We first define a generic temporal shift operation in transformers as follows:
	\begin{equation}
		\begin{aligned}
			\mathbf{Z}^t & = [ \mathbf{z}_0, \mathbf{z}_1,\dots,\mathbf{z}_{N} ], \\
			\mathbf{A} &= [\mathbf{a}_{0}, \mathbf{a}_{1}, \dots, \mathbf{a}_{N}], \\
			\mathbf{\hat{Z}}^t & = \mathbf{A}\odot \mathbf{Z}^{t'} + (\mathbf{1}-\mathbf{A}) \odot \mathbf{Z}^t,  
			\label{eq:shift_function}
		\end{aligned}
	\end{equation}
	
	\noindent 
	where $\mathbf{Z}^{t}, \mathbf{Z}^{t'}  \in \mathbb{R}^{D\times N}$ represent the patch features for current frame $t$ and another frame $t'$, respectively. $N$ is the number of patches, and $\mathbf{A}$ represents the matrix of shift channels, with $\mathbf{a}_{i} \in \mathbb{R}^{D}$ represents the vector of channel shifts for patch $i$, each element of which is equal to 0 or 1. $\mathbf{\hat{Z}}^t$ is the output image patches after shift operation.
	
	TSM~\cite{lin2019tsm} uses space-invariant channel shift for temporal redundancy modeling, which is a special case of our proposed patch shift operation by shifting $ \alpha $ percent of channels (with $\alpha$ percent of elements in $\mathbf{a}_i$ equal to 1), where $\alpha$ is a constant for all patches. In our case, we mainly explore temporal spatial mixing and shift patches in a space-variant manner, where $\mathbf{a}_i = \mathbf{0}$ or $\mathbf{1}$. To reduce the mixing space, shift pattern $\mathbf{p}$ is introduced, which is applied repeatedly in a sliding window manner to cover all patches. For example, $\mathbf{p} = \{0,1\}$ means shifting one patch for every two patches, therefore $\mathbf{A}=[\mathbf{0},\mathbf{1},\mathbf{0},\mathbf{1},\dots]$ in Eq.~\ref{eq:shift_function}. In practice, 2D shift patterns are designed for video data, which will be discussed in detail in the section below.

	\textbf{Patch shift SA.} By using the proposed patch shift operation, we can turn spatial-only SA into spatiotemporal SA. Given video patches $\mathbf{Z} \in \mathbb{R}^{D\times T\times N}$, the PatchShift function shifts the patches of each frame along the temporal dimension with pattern $\mathbf{p}$. As only part of patches are shifted in each frame, patches from different frames could be presented in the current frame, therefore, spatial-only SA naturally turns into a sparse spatiotemporal SA. After SA, patches from different frames are shifted back to their original locations. We follow~\cite{liu2021swin} to add a relative position bias with an extension to 3D position. To keep the track of shifted patches, the 3D positions are shifted alongside. With PatchShift, the multi-head SA is computed as:

	\begin{equation}
		\begin{aligned}
			\{ \mathbf{i}', \mathbf{Z}_{l-1}^{'} \} & =\text{PatchShift}(\mathbf{p}, \mathbf{i}, \mathbf{Z}_{l-1}), \\
			Q_{l}, K_{l}, V_{l}  & = W^{Q}_{l}\mathbf{Z}^{'}_{l-1},  W^{K}_{l}\mathbf{Z}^{'}_{l-1},  W^{V}_{l}\mathbf{Z}^{'}_{l-1} ,\\
			\mathbf{\hat{Z}} &	=\text{ShiftBack}( \text{SoftMax}(Q_{l}K_{l}^{T}/\sqrt{d} + B( \mathbf{i}'))V_{l}), 
			\label{eq:shift}
		\end{aligned}
	\end{equation}
	
	\noindent where $\{ \mathbf{i},\mathbf{Z}_{l-1} \}$ and $\{ \mathbf{i'}, \mathbf{Z}^{'}_{l-1}\}$ represent relative position bias indices and patches before and after PatchShift; $B$ is the bias matrix.

	\begin{figure}[t]
		\begin{minipage}[h]{0.5\textwidth} 
			\begin{center}
				\includegraphics[width=1\linewidth]{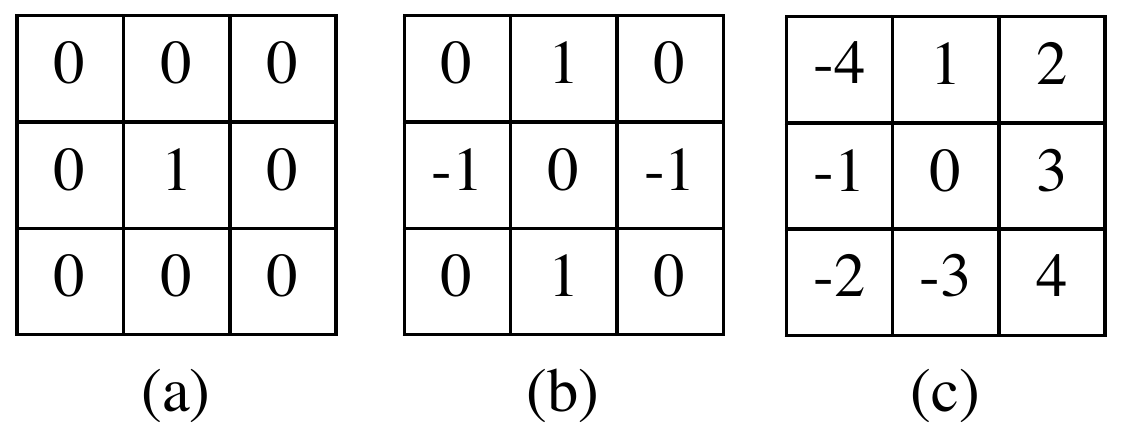}
			\end{center}
			\caption{Examples of patch shift patterns when patch number is $3\times 3$.}\label{fig:shift_pattern}
		\end{minipage}
		\hfill
		\begin{minipage}[h]{0.5\textwidth} 
			\begin{center}
				\tabcaption{Comparison of the complexities of different SA models.\\}
				\label{table:complexity}
				\begin{tabular}{c|c} \hline
					Attention & SA-Complexity \\ \hline
					Joint &  $  \mathcal {O}(N^2  T^2)$ \\
					Divide & $  \mathcal {O}(N^2  T + T^2 N)$  \\
					Sparse/Local & $   \mathcal {O}(\alpha N^2  T^2) $  \\
					PatchShift & $\mathcal {O}(N^2  T)$ \\
					\hline
				\end{tabular}
			\end{center}
		\end{minipage}
	\end{figure}

	\textbf{Patch shift patterns.} As mentioned before, in order to reduce the design space, our strategy is to employ repeated shift patterns, as it can scale up to different input sizes and is easy to implement. We adopt the following pattern design principles: a) Even spatial distribution. For each pattern, we uniformly sample the patches from the same frame to ensure they are evenly distributed. b) Reasonably large temporal receptive field. The temporal receptive field is set large enough to aggregate more temporal information. c) Adequate shift percentage. A higher percentage of shifted patches could encourage more inter-frame information communication. Various spatiotemporal SA models can be implemented with different patch shift patterns. We show several instantiations of $\mathbf{p}$ in Fig.~\ref{fig:shift_pattern}. The numbers represent the indices of the frames that the patches are from, where ``0", ``-" and ``+" indicate current, previous and next frames, respectively. Pattern (a) shifts a single patch from next frame to the center of current frame. Pattern (b) shifts patches with a ``Bayer filter" like pattern from previous and next frames. Pattern (c) shifts patches with a temporal field of 9, with patch from current frame in the center and patches from previous and next 4 frames around it. For window size larger than the pattern size, we spatially repeat the pattern to cover all patches. We use cyclic padding in~\cite{liu2021swin} for patches that exceed the temporal boundary. In the experiment section, we will discuss the design of shift patterns by extensive ablation studies.
	
	Patch shift is an efficient spatiotemporal SA modeling method for transformers as it only costs $\mathcal {O}(N^2 T)$ complexity in both computation and memory, which is much less than ``Joint" space-temporal SA, where $N$ and $T$ are the spatial and temporal dimension of patches. Patch shift is also more efficient than other factorized attention methods such as ``Divide"~\cite{GedasBertasius2021IsSA,AnuragArnab2021ViViTAV} (apply spatial-only and then temporal-only SA) and ``Sparse/Local"~\cite{GedasBertasius2021IsSA,ZeLiu2021VideoST} (subsample in space or temporal dimension). The complexity comparison of different SA models are in Table~\ref{table:complexity}.

	\begin{figure}[t]
		\setlength{\abovecaptionskip}{-1.mm}
		\setlength{\belowcaptionskip}{-1.mm}
		\begin{center}
			\includegraphics[width=0.7\textwidth,height=0.38\textwidth]{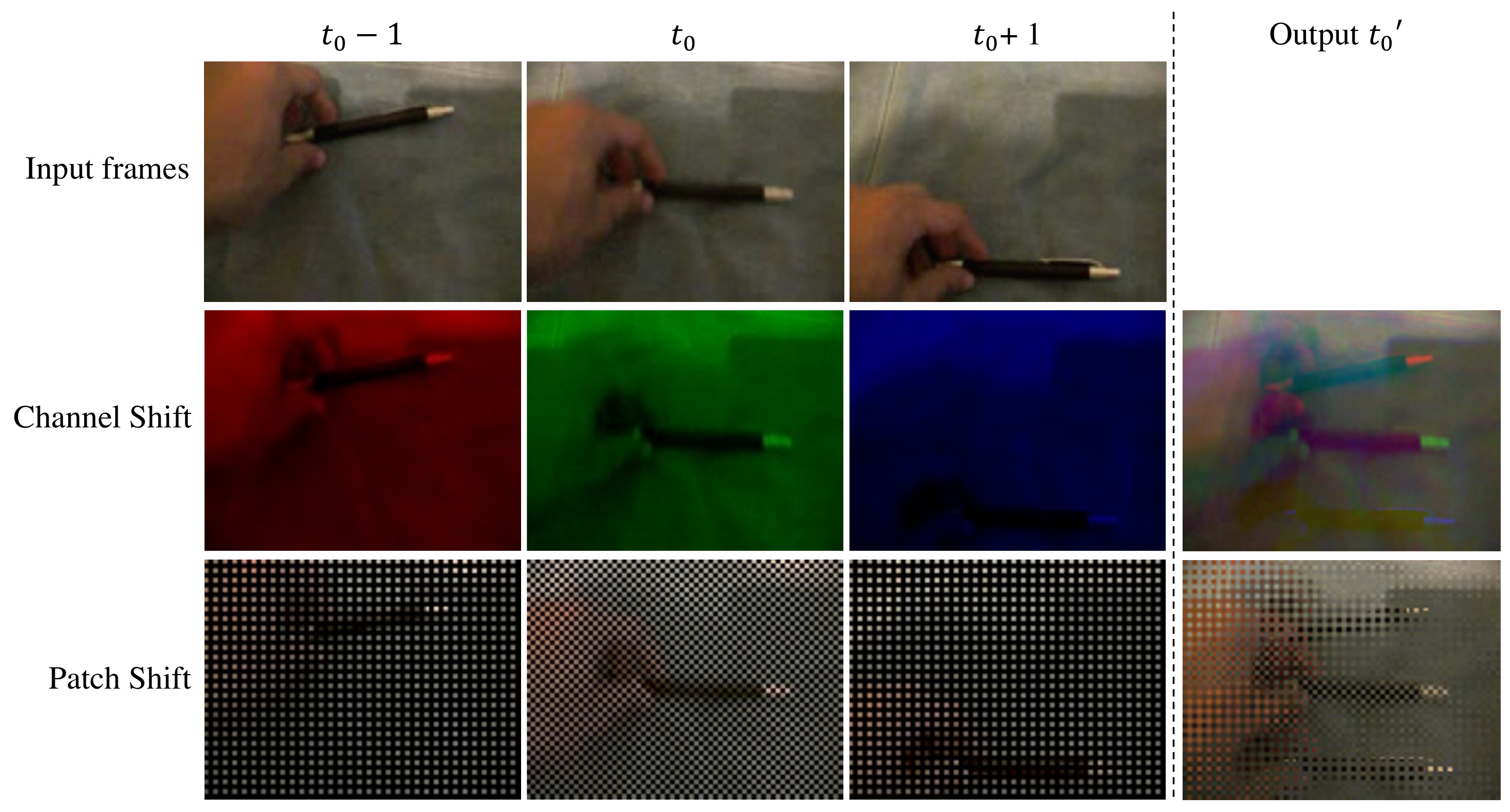}		
			\caption{An example of patch shift and channel shift for consecutive frames.}
			\label{fig:shift_vis}
		\end{center}
		
	\end{figure}

	\textbf{Discussions on patch and channel shifts.} Patch shift and channel shift are two zero parameter and low-cost temporal modeling methods. Patch shift is space-wise sparse and channel-wise dense, while channel shift is opposite. We show an example by applying patch shift and channel shift on three consecutive frames in Fig.~\ref{fig:shift_vis}. Here the shift operations are applied directly on RGB images for visualization, while in practice we apply shift operations on feature maps. In the output image of patch shifting, $1/4$, $1/2$ and $1/4$ patches are from frames $t_0 - 1$, $t_0$ and $t_0 + 1$, respectively. For channel shift, the red, green and blue colors represent frames $t_0 - 1$,  $t_0$ and $t_0 + 1$, respectively. The output of shift operation for frame $t_0$ is represented as $t_{0}'$ in the last column. As we can see from the figure, both patch shift and channel shift can capture the motion of action. Patch shift is spatially sparse while keeping the global channel information for each patch. In contrast, channel shift uses partial channel information in exchange for temporal information from other frames. Previous studies on vision transformer~\cite{chen2021crossvit} have shown that feature channels encode activation of different patterns or objects. Therefore, replacing partial feature channels of the current frame with other frames could potentially lose important information of patterns/objects of interest. In comparison, patch shift contains full information of channels of patches. When patch shift is employed for SA modeling, it builds a sparse spatiotemporal SA with 3D relations among patches. In comparison, channel shift can be viewed as a ``mix-patch" operation, by which temporal information is fused in each patch with shared 2D SA weights. Patch shift and channel shift perform shifting operations in orthogonal directions and they are complementary in nature.

	\subsection{Patch Shift Transformer} 
	
\begin{figure}[t]
		\setlength{\abovecaptionskip}{-1.mm}
		\setlength{\belowcaptionskip}{-1.mm}
		\begin{center}
			\includegraphics[width=1.0\linewidth]{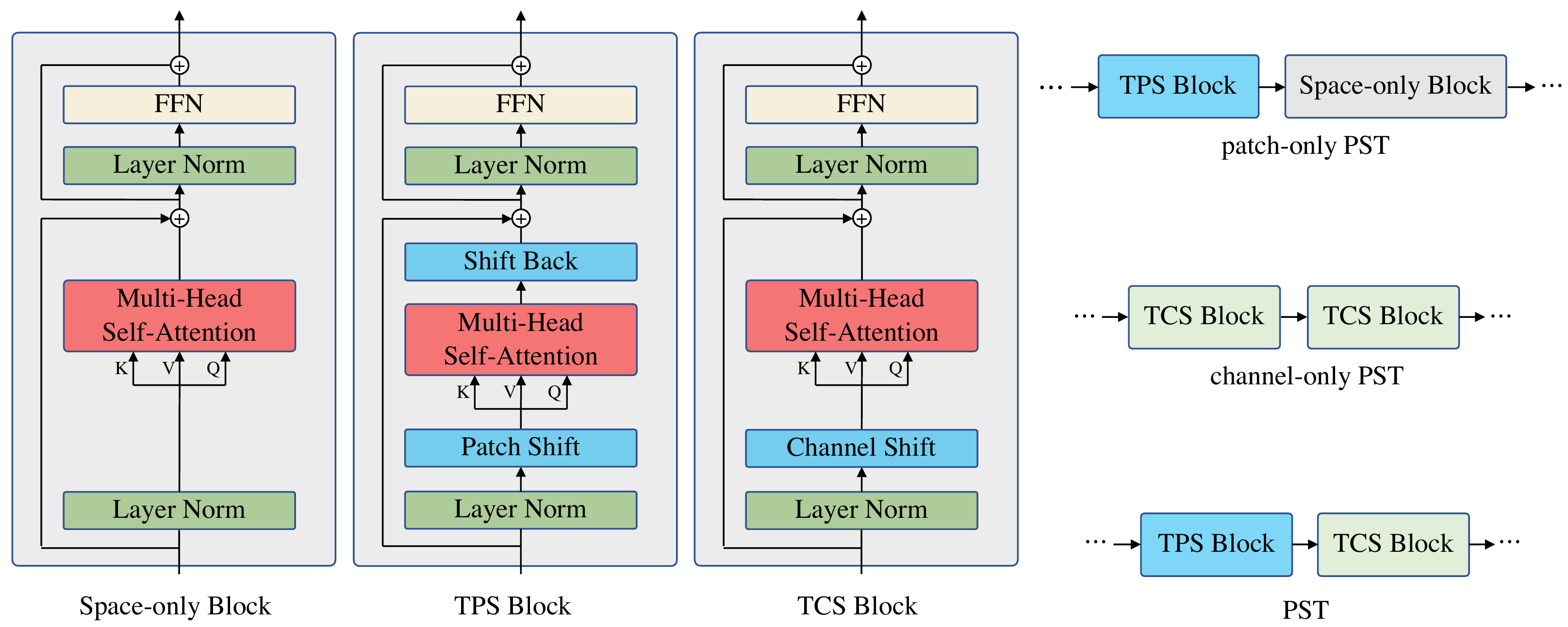}
		\end{center}
		\caption{
			An overview of building blocks and variants of PST.}\label{fig:variants}
		
	\end{figure}

	Based on the proposed TPS, we can build Patch Shift Transformers (PST) for efficient and effective spatiotemporal feature learning. A PST can be built by inserting TPS into the SA module of off-the-shelf 2D transformer blocks. Therefore, our model could directly benefit from the pre-trained models on 2D recognition tasks. The details of Temporal Patch Shift blocks (TPS block for short) can be seen in Fig.~\ref{fig:variants}. TPS turns spatial-only SA to spatiotemporal SA by aggregating information of patches from other temporal frames. However, it gathers information in a sparse manner and sacrifices SA within frames. To alleviate this problem, we insert one TPS block for every two SA modules (\textit{alternative shift} in short) so that spatial-only SA and spatiotemporal SA could work in turns to approximate full spatiotemporal SA. 
	
	We further improve the temporal modeling ability of spatial-only SA with channel shift. Specifically, partial channels of each patch are replaced with those from previous or next frames. We call this block a temporal channel shift (TCS) block. The final PST consists of both TPS and TCS blocks. We will also implement a channel-only PST for comparison to unveil the benefits of patch shift.
	
	\section{Experiments}

	\subsection{Dataset}
	
	\textbf{Something-Something V1$\&$V2}~\cite{goyal2017something} are large collections of video clips, containing daily actions interacting with common objects. They focus on object motion without differentiating manipulated objects. V1 includes 108,499 video clips, while V2 includes 220,847 video clips. Both V1 and V2 have 174 classes. \textbf{Kinetics400}~\cite{kay2017kinetics} is a large-scale dataset in action recognition, which contains 400 human action classes, with at least 400 video clips for each class. Each clip is collected from a YouTube video and then trimmed to around 10s. \textbf{Diving-48 V2}~\cite{li2018resound} is a fine-grained video dataset of competitive diving, consisting of 18k trimmed video clips of 48 unambiguous dive sequences. This dataset is a challenging task for modern action recognition systems as it reduces background bias and requires modeling of long-term temporal dynamics.  We use the manually cleaned V2 annotations of this dataset, which contains 15,027 videos for training and 1,970 videos for testing.

	\subsection{Experiment setup}
	
	\textbf{Models.} We choose swin transformer~\cite{liu2021swin} as our backbone network and develop PST-T, PST-B with an increase in model size and FLOPs based on swin transformer Tiny and Base backbones, respectively. In ablation study, we use Swin-Tiny considering its good trade-off between performance and efficiency. 
	We adopt 32 frames as input and the tubelet embedding strategy in ViViT~\cite{AnuragArnab2021ViViTAV} with patch size $2\times4\times4$ by default. As PST-T and PST-B are efficient models, when comparing with SOTA methods, we introduce PST-T$\dag$ and PST-B$\dag$, which doubles the temporal attention window to 2 with slightly increased computation.

	\textbf{Training.} For all the datasets, we first resize the short side of raw images to $256$ and then apply center cropping of $224\times224$. During training, we follow~\cite{ZeLiu2021VideoST} and use random flip,  AutoAugment~\cite{autoaugment} for augmentation. We utilize AdamW~\cite{LoshchilovH19} with the cosine learning rate schedule for network training. For PST-T, the base learning rate, warmup epoch, total epoch, stochastic depth rate, weight decay, batchsize are set to $10^{-3}$, 2.5, 30, 0.1, 0.02, 64 respectively. For larger model PST-B, learning rate, drop path rate and weight decay are set to $3\times10^{-4}$, 0.2, 0.05, respectively. 
	
	\textbf{Testing.} For fair comparison, we follow the testing strategy in previous state-of-the-art methods. We report the results of two different sampling strategies. On Something-something V1$\&$V2 and Diving-48 V2, uniform sampling and center-crop (or three-crop) testing are adopted. On Kinetics400, we adopt the dense sampling strategy as in~\cite{AnuragArnab2021ViViTAV} with 4 view, three-crop testing.
	
	\subsection{Ablation study}
	
	To investigate the design of patch shift patterns and the use of TPS blocks, we conduct a series of experiments in this section. All the experiments are conducted on Something-something V1 with Swin-Tiny as backbone (IN-1K pretrained). For experiments on design of patch shift we use patch-only PST for clarity. 
	
	\textbf{The number and distribution of shifted patches.}	We start with a simple experiment by shifting only one center patch along temporal dimension within each window. It can be seen from Table~\ref{tab:ablation_table}(a) that this simple shift pattern (center-one) brings significant improvements ($4.7\%$ on top-1) over the model without shifting operation (none). We then increase the number of shifting patches to $1/2$ of the total patches, however, in an uneven distribution (shift only the left half of patches). This uneven shift pattern does not improve over the simple ``center-one" pattern. However, when the shifted patches are distributed evenly within the window (even-2), the performance increases by $0.9\%$ on top-1. This indicates that, the shifted and non-shifted patches should be distributed evenly. It is also found that a large temporal field is helpful. When we increase the temporal field to 3 by shifting 1/4 patches to previous frame and 1/4 patches to next frame (even-3), the performance is improved by $2.4\%$ on top-1 over shifting patches in one dimension.

	\begin{table}[h]
		\setlength{\abovecaptionskip}{-1.mm}
		\setlength{\belowcaptionskip}{-1.mm}
		\caption{Ablation studies on TPS. All the experiments are conducted on Something-something V1 with Swin-Tiny as backbone.}
		\label{tab:ablation_table}
		\subfigure[\begin{scriptsize}{Patch distribution}\end{scriptsize}]{%
			\setlength{\tabcolsep}{1.7mm}
			\resizebox{0.31\textwidth}{12mm}{
				\begin{tabular}{c|cc}
					\toprule[1.0pt]
					Distribution & Top-1  & Top-5 \\
					\hline
					
					None  & 40.6 & 71.4 \\
					Center-one & 45.3 &  75.1   \\
					Uneven & 45.3  & 75.5  \\
					Even-2 & 46.2 & 76.1    \\
					Even-3 & \textbf{48.6} & \textbf{77.8}    \\
					\bottomrule[1.0pt]
			\end{tabular}}\hfill}\hfill
		\subfigure[\begin{scriptsize}{Shift patterns}\end{scriptsize}]{%
			\setlength{\tabcolsep}{1.7mm}
			\resizebox{0.28\textwidth}{10.5mm}{
				\begin{tabular}{c|cc}
					\toprule[1.0pt]
					Pattern &  Top-1  & Top-5 \\
					\hline
					
					A-3 & 48.6 &  77.8   \\
					B-4 & 50.7 &  79.3   \\
					C-9 & \textbf{51.8} &  \textbf{80.3}   \\
					D-16 &50.0 & 79.5   \\
					\bottomrule[1.0pt]
			\end{tabular}}\hfill}\hfill
		\subfigure[\begin{scriptsize}{Number of stages with TPS }\end{scriptsize}]
		{%
			\setlength{\tabcolsep}{1.3mm}
			\resizebox{0.31\textwidth}{12mm}{
				\begin{tabular}{cccc|cc}
					\toprule[1.0pt]
					\multicolumn{4}{c|}{Stage}  & \multirow{2}{*}{Top-1}  & \multirow{2}{*}{Top-5} \\ 
					
					1 & 2 & 3 &4 & & \\
					\hline
					$\checkmark$ &  &  &  & 47.3 &  77.0   \\
					$\checkmark$ & $\checkmark$ &  & &  48.4 & 77.6    \\
					$\checkmark$ & $\checkmark$  & $\checkmark$ &  &50.4  & 79.1   \\
					$\checkmark$ & $\checkmark$ & $\checkmark$ &$\checkmark$  & \textbf{51.8} & \textbf{80.3}  \\
					
					\bottomrule[1.0pt]
			\end{tabular}}\hfill
		}\hfill
		
		\subfigure[\begin{scriptsize}{Shift back, Alternative shift and shift RPE}\end{scriptsize}]{%
			\setlength{\tabcolsep}{1.2mm}
			\resizebox{0.48\textwidth}{10mm}{
				\begin{tabular}{c|c|c|cc}
					\toprule[1.0pt]
					Shift back & Alternative & Shift RPE  & Top-1 & Top-5 \\ 
					
					\hline
					& $\checkmark$ & $\checkmark$ &  47.3  & 77.0  \\
					$\checkmark$ &  & $\checkmark$  &  46.4& 76.6  \\
					$\checkmark$ & $\checkmark$  &  &  46.1 &  76.0   \\
					$\checkmark$ & $\checkmark$ & $\checkmark$   & \textbf{51.8} & \textbf{80.3}  \\
					
					\bottomrule[1.0pt]
			\end{tabular}}\hfill
		}\hfill
		\subfigure[\begin{scriptsize}{Comparison of spatiotemporal attentions}\end{scriptsize}]{%
			\setlength{\tabcolsep}{1.5mm}
			\resizebox{0.46\textwidth}{13mm}{
				\begin{tabular}{c|c|c|c|cc}
					\toprule[1.0pt]
					& FLOPs & Memory & Top-1  & Top-5 \\
					\hline
					Avgpool  & 72G  &  3.7G & 40.6 & 71.4 \\
					Joint & 106G  & 20.2G &  51.5 & 80.0  \\
					Local & 88G  & 11G & 49.9 & 79.2 \\
					Sparse & 72G  &  4.0G  &  42.7  & 74.0  \\ \hline
					Channel-only & 72G &  3.7G & 51.2 & 79.7 \\
					Patch-only & 72G & 3.7G & 51.8  & 80.3  \\
					PST & \textbf{72G} & \textbf{3.7G} & \textbf{52.2}  & \textbf{80.3}  \\
					\bottomrule[1.0pt]
			\end{tabular}}\hfill
		} \hfill

	\end{table}
	
	\textbf{Patch shift patterns.} Based on the experiments in Table~\ref{tab:ablation_table}(a), we design a few different patch shift patterns with various temporal fields. The patches of different frames are distributed evenly within the window. Pattern A with temporal field 3 is shown in Fig.~\ref{fig:shift_pattern}(b), which is ``Bayer filter" like. Pattern B with temporal field 4 is implemented by replacing $1/4$ of index ``0" frame patches in pattern A as index ``2" frame patches. Pattern C is shown in Fig.~\ref{fig:shift_pattern}(c). Pattern D is designed by placing patches from 16 consecutive frames in a $4\times 4$ pattern grid. More details can be found in the supplementary materials. We can see from the Table~\ref{tab:ablation_table}(b) that the performance of TPS gradually increases when the temporal field grows. The best performance is reached at temporal field 9, which achieves a good balance between spatial and temporal dimension. We use this pattern for the rest of our experiments.
	
	\textbf{The number of stages using TPS blocks.} As can be seen in Table~\ref{tab:ablation_table}(c), by using TPS blocks in more stages of the network, the performance gradually increases. The model achieves the best performance when all the stages are equipped with TPS blocks.
	
	\textbf{Shift back, alternative shift and shift RPE.} Shift back operation recovers the patches' locations and keeps the frame structure complete. Alternative shift is described in Section 3.3 for building connections between patches. Shift RPE represents whether relative positions are shifted alongside patches. As shown in Table~\ref{tab:ablation_table}(d), removing each of them would decrease performance by $4.5\%, 5.4\%, 5.7\%$, respectively. Therefore, we use all the operations in our model.

	\textbf{Comparison with other temporal modeling methods.} In Table~\ref{tab:ablation_table}(e), we compare PST with other designs of spatiotemporal attention methods in FLOPs, peak training memory consumption and accuracy. ``Avgpool" achieves only $40.6\%$ Top-1 rate since it cannot distinguish temporal ordering. Joint spatiotemporal SA achieves a good performance at the price of high computation and memory cost. Local spatiotemporal attention~\cite{ZeLiu2021VideoST} applies SA within a local window with 3D window size. It reduces the computation and memory cost but at the price of performance drop comparing to ``Joint". We also implement a sparse spatiotemporal SA by subsampling spatial patches to half, while keeping the full temporal dimension. It performs poorly as only part of patches participate in SA computation in each layer. We implement channel-only PST with shift ratio equals $1/4$, which is the same in the~\cite{lin2019tsm}. This channel-only PST also achieves strong performance. 
	
	Patch-only PST outperforms all other temporal modeling methods without additional parameters and FLOPs. The best performance comes from combining channel-only and patch-only PST, which indicates they are complementary in nature. Specifically, PST exceeds joint spatiotemporal SA with much less computation and only 1/5 of memory usage. It also outperforms other efficient spatiotemporal SA models with fewer computation and memory cost.
	
	\begin{table}[h]
		\begin{center}
			\caption{Comparisons with the other methods on Something-something V1 \& V2.}
			\label{table:sthv1}
			\resizebox{0.88\textwidth}{44mm}{
				\begin{tabular}{ccccccccccc}
					\hline
					\multirow{2}{*}{Model} & \multirow{2}{*}{Pretrain} & \multirow{2}{*}{Crops $\times$ Clips} & \multirow{2}{*}{FLOPs} & \multirow{2}{*}{Params} & \multicolumn{2}{c}{Sthv1} & \multicolumn{2}{c}{Sthv2} \\ 
					
					&  &  &  &  & Top-1 & Top-5 & Top-1 & Top-5 \\ \hline
					
					TSM~\cite{lin2019tsm} & K400  & $3\times2$ & 65G & 24.3M & - & - & 63.4 & 88.5  \\
					
					\multirow{1}{*}{TEINet~\cite{liu2020teinet}} & IN-1K & $1\times1$ & 66G & 30.4M  & 49.9 & - & 62.1 & -  \\ 
					
					\multirow{1}{*}{TEA~\cite{li2020tea}} &IN-1K & $1\times1$ & 70G  & 24.3M& 51.9 & 80.3 & - & -  \\

					TDN~\cite{wang2021tdn} & IN-1K & $1\times1$ & $72G$& 24.8M & 53.9 & 82.1 & 65.3 & 89.5  \\
					
					\multirow{1}{*}{ACTION-Net~\cite{Wang_2021_CVPR}} &  IN-1K &$ 1\times1$ & 70G & 28.1M & - & - & 64.0 & 89.3 \\

					SlowFast R101, 8x8~\cite{feichtenhofer2019slowfast} & K400 &  $ 3\times1 $ & 106G & 53.3M & -  & -  &  63.1 & 87.6 \\
					MSNet~\cite{kwon2020motionsqueeze} & IN-1K &  $ 1\times1 $ & 101G & 24.6M & 52.1 & 82.3 &  64.7 & 89.4 \\
					blVNet~\cite{fan2019blvnet} & IN-1K & $ 1\times1$ & $129G$& 40.2M & -  & - & 65.2  & 90.3   \\

					\hline 
					
					Timesformer-HR~\cite{GedasBertasius2021IsSA} & IN-21K & $ 3\times1$ & 1703G & 121.4M & - & - & 62.5  & -   \\

					ViViT-L/16x2~\cite{AnuragArnab2021ViViTAV} & IN-21K & $3\times1$ & 903G & 352.1M & - & - & 65.9  & 89.9   \\
					MViT-B, 64×3~\cite{Fan_2021_ICCV}  & K400 & $3\times1$ & 455G& 36.6M & - & - & 67.7  & 90.9   \\
					Mformer-L~\cite{patrick2021keeping} & K400 & $ 3\times1$ & 1185G& 86M & - & - & 68.1  & 91.2   \\
					X-ViT~\cite{bulat2021space} & IN-21K & $3\times1$ & 283G & 92M & - & - & 66.2  & 90.6   \\
					SIFAR-L~\cite{SIFAR}  & K400 & $ 3\times1$ & 576G & 196M & - & - & 64.2  & 88.4   \\
					Video-Swin~\cite{liu2021video} & K400 & $3\times1$ & 321G & 88.1M & - & - & 69.6  & 92.7   \\
					\hline 
					\multirow{4}{*}{PST-T} & IN-1K & $1\times1$ & \multirow{5}{*}{72G}& \multirow{5}{*}{28.5M} & 52.2 & 80.3 & 65.7 & 90.2   \\
					& IN-1K & $ 3\times1$ &   & & 52.8 & 80.5 & 66.4 & 90.2   \\
					& K400 & $ 1\times1$ &   &  & 53.2 & 82.2 & 66.7 & 90.6 \\
					& K400 & $ 3\times1$ &   &  & 53.6 & 82.2 & 67.3 &  90.5  \\ 
					\multirow{1}{*}{PST-T$\dag$} & K400 & $3\times1$ & 74G & & 54.0 & 82.3 & 67.9 & 90.8   \\
					\hline
					\multirow{4}{*}{PST-B} & IN-21K & $1\times1$ & \multirow{5}{*}{247G} & \multirow{5}{*}{88.8M} & 55.3 & 81.9 & 66.7 & 90.7 \\

					&  IN-21K &$3\times1$ & & & 55.6 & 82.2 & 67.4 & 90.9  \\
					&  K400 &$1\times1$ &  & & 57.4 & 83.2 &68.7  & 91.3   \\
					&  K400 &$3\times1$ &  & & 57.7 & 83.4 & 69.2 & 91.9  \\ 
					\multirow{1}{*}{PST-B$\dag$} & K400 & $3\times1$ & 252G &  & \textbf{58.3} & \textbf{83.9} & \textbf{69.8} & \textbf{93.0} \\
					\hline

					\hline

				\end{tabular}
			}
		\end{center}

	\end{table}

	\subsection{Comparison with SOTA}
	
	\textbf{Something V1 $\&$ V2.} The performance statistics on Something V1 $\&$ V2, including the pretrained dataset, classification results, inference protocols, the corresponding FLOPs and parameter numbers are shown in Table~\ref{table:sthv1}.
	
	The first compartment contains methods based on 3D CNNs or factorized (2+1)D CNNs. Using efficient inference protocol (16 views and center crop$\times$1 clip), ImageNet1K pretrain, PST-T obtains 52.2$\%$ accuracy on V1 and 65.7$\%$ on V2, respectively, which outperforms all the CNN-based methods with similar FLOPs. Note that TDN~\cite{wang2021tdn} uses a different sampling strategy comparing to other methods, which requires 5 times more frames. Even though, PST-T still outperforms TDN on larger Something-something V2.
	
	The second compartment contains transformer-based methods. Our small model PST-T pretrained on Kinetics400 offers competitive performance on SomethingV2 comparing to these methods. PST-B is a larger model and pretrained on ImageNet21K/Kinetics400. PST-B achieves $57.7\%$ and $69.2\%$ on V1 $\&$ V2, outperforming Timesformer-HR~\cite{GedasBertasius2021IsSA}, ViViT-L~\cite{AnuragArnab2021ViViTAV} and MViT-B~\cite{Fan_2021_ICCV} at a lower cost on computation or parameter. PST-B also outperforms Mformer-L~\cite{patrick2021keeping} and X-ViT~\cite{bulat2021space}, which are recently proposed efficient temporal modeling methods. They apply trajectory self-attention and local spatiotemporal self-attention, respectively. Our PST-B$\dag$ achieves $58.3\%$ and $69.8\%$ on V1 $\&$ V2, which outperforms other transformer-based methods. Note that, SIFAR-L~\cite{SIFAR} uses larger backbone network Swin-L, however, its performance is less satisfactory. PST-B$\dag$ also outperforms Video-Swin~\cite{ZeLiu2021VideoST}, which uses full temporal window on this dataset. The performances of PST family on Something-something V1$\&$V2 confirms its remarkable ability for spatiotemporal modeling.

	\textbf{Kinetics400.} We report our results on scene-focused Kinetics400 in Table~\ref{table:kinetics} and compare them with previous state-of-the-arts. As we can see from Table~\ref{table:kinetics}, PST-T achieves $78.2\%$ top-1 accuracy and outperforms majority of (2+1D) CNN-based methods such as TSM~\cite{lin2019tsm}, TEINet~\cite{liu2020teinet}, TEA~\cite{li2020tea}, TDN~\cite{wang2021tdn} with less total FLOPs. Our larger model PST-B achieves $81.8\%$ top-1 accuracy, which outperforms strong 3D-CNN counterparts such as SlowFast~\cite{feichtenhofer2019slowfast} and X3D~\cite{Feichtenhofer_2020_CVPR}. 
	
	Comparing to transformer-based methods such as Timesformer~\cite{GedasBertasius2021IsSA} and ViViT-L~\cite{AnuragArnab2021ViViTAV}, our PST-B$\dag$ achieves $82.5 \%$ with less computation overheads. Specifically, PST-B$\dag$ outperforms SIFAR-L~\cite{SIFAR} which adopts larger size Swin-L as backbone network. PST-B$\dag$ also achieves on par performance with recently developed Video-Swin~\cite{liu2021video} with less computation cost.

	\begin{table}[h]

		\begin{center}
			\caption{Comparisons with the state-of-the-art methods on Kinetics400.}
			\label{table:kinetics}
			\resizebox{0.7\textwidth}{38mm}{
				\begin{tabular}{ccccccccc}
					\hline
					Model & Pretrain &  Crops $\times$ Clips & FLOPs & Params & Top-1 & Top-5 \\ \hline
					
					I3D~\cite{carreira2017quo} & IN-1K & $ 1\times1$ & 108G& 28.0M & 72.1 & 90.3  \\
					NL-I3D~\cite{NonLocal2018} &IN-1K & $6\times10$ & 32G& 35.3M & 77.7 & 93.3 \\
					CoST~\cite{Li_2019_CVPR} &IN-1K & $3\times10$ & 33G & 35.3M & 77.5 & 93.2 \\
					SlowFast-R50~\cite{feichtenhofer2019slowfast} &IN-1K & $ 3 \times 10$ & 36G  & 32.4M   & 75.6 & 92.1 \\
					
					X3D-XL~\cite{Feichtenhofer_2020_CVPR} & - & $ 3\times10$ & 48G  &  11.0M  & 79.1 & 93.9 \\
					
					\multirow{1}{*}{TSM~\cite{lin2019tsm}} & IN-1K& $ 3\times10$ & 65G & 24.3M & 74.7 & 91.4  \\
					
					TEINet~\cite{liu2020teinet} & IN-1K & $3\times10$  & 66G  & 30.4M & 76.2 & 92.5  \\
					
					TEA~\cite{li2020tea} & IN-1K & $ 3\times10$  &  70G  & 24.3M & 76.1 & 92.5  \\
					
					\multirow{1}{*}{TDN~\cite{wang2021tdn}} & IN-1K & $3\times10$& $72G $ & 24.8M & 77.5 & 93.2  \\

					\hline 
					
					Timesformer-L~\cite{GedasBertasius2021IsSA} & IN-21K & $ 3\times1$ & 2380G& 121.4M & 80.7  & 94.7  \\
					
					ViViT-L/16x2~\cite{AnuragArnab2021ViViTAV} & IN-21K & $ 3\times1$ & 3980G & 310.8M & 81.7 & 93.8   \\
					X-ViT~\cite{bulat2021space} & IN-21K & $ 3\times1$ & 283G & 92M & 80.2  & 94.7  \\
					MViT-B, 32×3~\cite{Fan_2021_ICCV}  & IN-21K & $ 1\times5$ & 170G& 36.6M & 80.2  & 94.4   \\
					MViT-B, 64×3~\cite{Fan_2021_ICCV}  & IN-21K & $ 3\times3$ & 455G & 36.6M & 81.2  & 95.1   \\
					Mformer-HR~\cite{patrick2021keeping}  & K400 & $ 3\times1$ & 959G& 86M & 81.1  & 95.2   \\
					TokenShift-HR~\cite{HaoZhang2021TokenST} & IN-21K & $ 3\times10$ & 2096G & 303.4M &  80.4 & 94.5   \\
					SIFAR-L~\cite{SIFAR}  & IN-21K & $ 3\times1$ &  576G& 196M &  \textbf{82.2} & \textbf{95.1}   \\	
					Video-Swin~\cite{ZeLiu2021VideoST}  & IN-21K & $ 3\times4$ & 282G& 88.1M &  \textbf{82.7} & \textbf{95.5}  \\	
					\hline
					
					\multirow{1}{*}{PST-T} & IN-1K & $3\times4$ & 72G & \multirow{1}{*}{28.5M}  & 78.2 & 92.2  \\
					\multirow{1}{*}{PST-T$\dag$} & IN-1K & $3\times4$ & 74G& \multirow{1}{*}{28.5M}  & 78.6 & 93.5  \\
					
					\multirow{1}{*}{PST-B} & IN-21K & $3\times4$ & 247G& \multirow{1}{*}{88.8M} &  81.8 & 95.4  \\
					\multirow{1}{*}{PST-B$\dag$} & IN-21K & $3\times4$ & 252G & \multirow{1}{*}{88.8M} &  \textbf{82.5} & \textbf{95.6}  \\

					\hline

				\end{tabular}
			}
		\end{center}
		
	\end{table}

	\textbf{Diving-48 V2.} In Table~\ref{table:diving48}, we further evaluate our method on fine-grained action dataset Diving-48. As older version of diving-48 has labeling errors, we compare our methods with reproduced SlowFast~\cite{feichtenhofer2019slowfast} and Timesformer in~\cite{GedasBertasius2021IsSA} on V2. PST-T achieves competitive $79.2\%$ top-1 performance with ImageNet-1K pretrain and costs much less computation resources than SlowFast, Timersformer and Timesformer-HR. When PST-T is pretrained on Kinetics400, it also outperforms Timersformer-L. Our larger model PST-B outperforms TimsFormer-L by a large margin of $2.6\%$ when both are pretrained on ImageNet-21K with 9.7$\times$ less FLOPs. When model is pretrained on Kinetic400, PST-B achieves $85.0\%$ top-1 accuracy. PST-T$\dag$ and PST-B$\dag$ further boost the performance to $82.1\%$ and $86.0\%$, respectively. 
	
	\begin{table}[h]
		\begin{center}
			\caption{Comparisons with the other methods on Diving-48 V2.}
			\label{table:diving48}
			\resizebox{0.68\textwidth}{18mm}{
				\begin{tabular}{ccccccccc}
					\hline
					Model & Pretrain & Crops $\times$ Clips & FLOPs & Params & Top-1  & Top-5   \\  \hline
					
					SlowFast R101, 8x8~\cite{feichtenhofer2019slowfast} & K400 &  $ 3\times1 $ & 106G & 53.3M & 77.6 & - \\ \hline
					Timesformer~\cite{GedasBertasius2021IsSA} & IN-21K & $ 3\times1$ & 196G& 121.4M & 74.9  & -   \\			
					Timesformer-HR~\cite{GedasBertasius2021IsSA} & IN-21K & $ 3\times1$ & 1703G& 121.4M & 78.0  & -   \\
					Timesformer-L~\cite{GedasBertasius2021IsSA} & IN-21K & $ 3\times1$ & 2380G& 121.4M & 81.0  & -   \\

					\hline 
					\multirow{2}{*}{PST-T} & IN-1K & $ 3\times1$ & 72G  & \multirow{3}{*}{28.5M} & 79.2 & 98.2    \\
					& K400 & $ 3\times1$ & 72G  &  & 81.2 & 98.7   \\
					\multirow{1}{*}{PST-T$\dag$} &  K400 &$ 3\times1$ & 74G &   & 82.1 & 98.6   \\
					\hline
					
					\multirow{2}{*}{PST-B} &  IN-21K &$ 3\times1$ & 247G & \multirow{3}{*}{88.1M} & 83.6 & 98.5   \\
					
					&  K400 &$ 3\times1$ & $247G$ & & 85.0 & 98.6   \\
					\multirow{1}{*}{PST-B$\dag$} &  K400 &$ 3\times1$ & 252G &   & \textbf{86.0} & \textbf{98.6}   \\
					\hline

				\end{tabular}
			}
		\end{center}
		
	\end{table}
	
	\textbf{Latency, throughput and memory.} The inference latency, throughput (videos/second) and inference memory consumption are important for action recognition in practice. We performed measurement on a single NVIDIA Tesla V100 GPU. We use batch size of 1 for latency measurement and batch size of 4 for throughput measurement. The accuracy is tested on validation set of Something-something V1\&V2 dataset. The results are shown in Table~\ref{table:latency_memory}. We make three observations: (1) Comparing to baseline ``2D Swin-T" that uses 2D transformer with avgpooling for temporal modeling, PST-T has slightly higher latency, same memory consumption, but much better performance ($11.6\%$, $9.0\%$ top-1 improvements on Sthv1\&v2, respectively). (2) Comparing to ``Video-Swin-T" that utilizes full scale temporal information, PST-T uses $76\%$ less memory and has 2 times faster inference speed, while achieves very competitive performance. (3) For larger model PST-B$\dag$, it achieves better performance than Video-Swin-B while using $50\%$ less inference memory and $80\%$ less time.
	
	\begin{table}
		\setlength{\abovecaptionskip}{-1.mm}
		\setlength{\belowcaptionskip}{-1.mm}  
		\begin{center}
			\caption{Memory and latency comparison on Something-something V1\&V2 (Measured on NVIDIA Tesla V100 GPU)}
			\label{table:latency_memory}
			\setlength{\tabcolsep}{1.5mm}
			\resizebox{1\textwidth}{14mm}{
				\begin{tabular}{c|c|c|c|c|c|cccc}
					\toprule[1.0pt]
					\multirow{2}{*}{Methods} & \multirow{2}{*}{FLOPs} & \multirow{2}{*}{Param} &  \multirow{2}{*}{Memory} & \multirow{2}{*}{Latency} & \multirow{2}{*}{Throughput} &
					\multicolumn{2}{c}{Sthv1} & \multicolumn{2}{c}{Sthv2} \\ 
					
					&  &  & & &  & Top-1 & Top-5 & Top-1 & Top-5 \\ \hline

					2D Swin-T  & 72G &   & 1.7G & 29ms & 35.5 v/s  & 40.6 & 71.4 & 56.7 & 84.1  \\
					
					Video-Swin-T~\cite{ZeLiu2021VideoST} &  106G($\uparrow34$G) & 28.5M & 3.0G($\uparrow1.3$G) & 62ms($\uparrow33$ms) & 17.7 v/s & 51.5 & 80.0 & 65.7 & 90.1 \\
					
					PST-T &  72G &  & 1.7G & 31ms($\uparrow2$ms) & 34.7 v/s & \textbf{52.2} & \textbf{80.3} & \textbf{65.7} & \textbf{90.2} \\ \hline
					
					2D Swin-B  & 247G &  & 2.2G & 71ms &  15.5 v/s  & - & - & 59.5 & 86.3 \\
					
					Video-Swin-B~\cite{ZeLiu2021VideoST} & 321G($\uparrow74$G)  &  88.8M  & 3.6G($\uparrow1.4$G) & 147ms($\uparrow76$ms) & 7.9 v/s & - & - & 69.6 & 92.7  \\
					
					PST-B$\dag$ &  252G($\uparrow5$G) &  & 2.4G($\uparrow0.2$G) & 81ms($\uparrow 10 $ms) & 13.8 v/s & - & - & \textbf{69.8}  & \textbf{93.0}  \\
					
					\bottomrule[1.0pt]
			\end{tabular}}\hfill
		\end{center}
	\end{table}

	\section{Conclusions}
	
	In this paper, we proposed a novel temporal patch shift method, which can be inserted into transformer blocks in a plug-and-play manner for efficient and effective spatiotemporal modeling. By comparing the patch shift and channel shift operations under transformer framework, we showed that they are two efficient temporal modeling methods and complementary in nature. PST achieved competitive performance comparing to previous methods on the datasets of Something-something V1\&V2 ~\cite{goyal2017something}, Diving-48~\cite{li2018resound} and Kinetics400~\cite{kay2017kinetics}. In addition, we presented in-depth ablation studies and visualization analysis on PST to investigate the design principles of patch shift patterns, and explained why it is effective for learning spatiotemporal relations. PST achieved a good balance between accuracy and computational cost for effective action recognition.
	
	\appendix
	
	\section{Appendix}
	
	\subsection{More details on the implementation of TPS}
	
	We present the ``PyTorch style" pseudo-code of a special case of TPS, where the patches from neighboring frames are shifted using a ``Bayer filter" pattern. The patch shift module is placed before the self-attention calculation, and then a shift back operation is employed after the self-attention calculation.
	\begin{lstlisting}
# Patch shift with pattern B
def PatchShift(x, shift_back):
	B, C, T, H, W = x.size()
	direct = 1
	if shift_back:
		direct = -1
	x[:,:,:,0::2,1::2] = roll(x[:,:,:,0::2,1::2], shifts= direct, dims=2)
	x[:,:,:,1::2,0::2] = roll(x[:,:,:,1::2,0::2], shifts=-direct, dims=2)
	return x
		
		
	\end{lstlisting}
	
	The simple patch shift function described above transfers a vanilla spatial self-attention to a spatiotemporal self-attention. Other patterns can be implemented in a similar paradigm. As we use Swin Transformer~\cite{liu2021swin} as our backbone, the coordinates of patches are shifted alongside the patches. 
	
	\subsection{More details on the shift patterns}
	We show the Patterns A, B, C and D in Figure~\ref{fig:shift_pattern} here. Pattern B is obtained via replacing 1/4 index-0 frame patches in pattern A by index-2 frame patches. Pattern C is obtained by placing patches from 9 consecutive frames in a 3 $\times$ 3 grid. Pattern D is obtained by placing patches from 16 consecutive frames in a 4 $\times$ 4 grid.

	\begin{figure}[h]
		\vspace{-1mm}
		\begin{center}
			\includegraphics[width=0.5\textwidth]{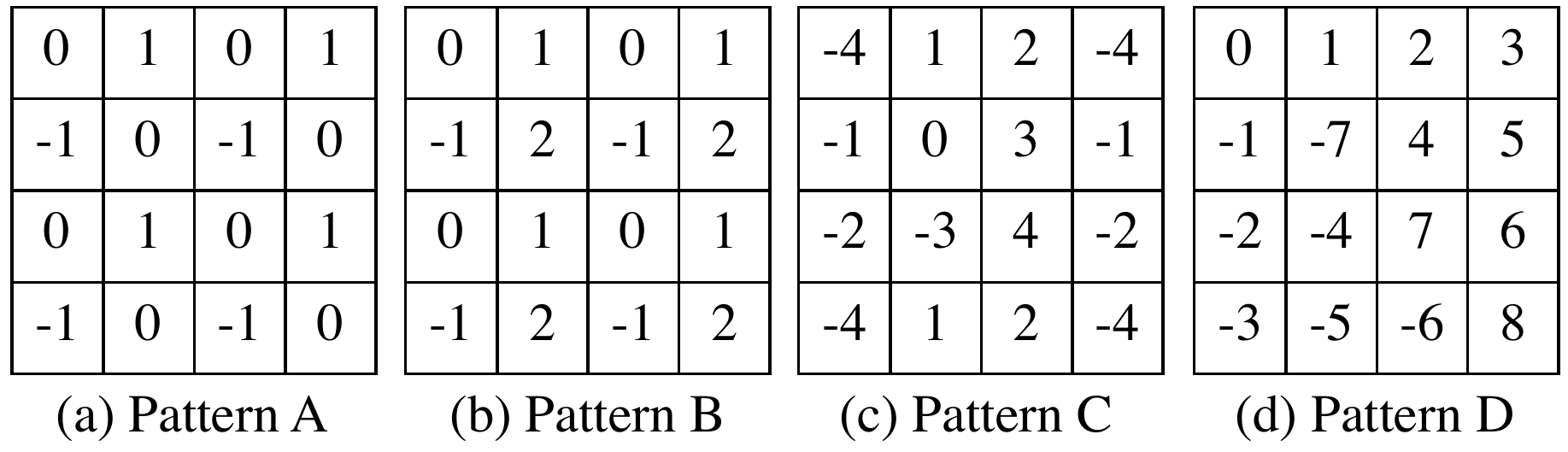}
			\caption{Pattern A, B, C and D in Table 2(b) of the main paper.}
			\label{fig:shift_pattern}
		\end{center}
		
	\end{figure}

	\subsection{Additional results on DeiT backbone}
	
	
	TPS is a plug-and-play module and it can be embedded into many existing transformers to increase their spatiotemporal modeling capabilities. To illustrate the effectiveness of TPS, we further test our method on another popular backbone DeiT~\cite{touvron2021training}. We use DeiT-S for our experiment, which has $14\times14$ image patches at every layer and an additional class token.  We insert a TPS module in every two blocks of DeiT and operate directly on image patches. We densely sample 16 frames as input for training. Other training and testing configurations are the same as PST-T. 
	
	We compare our proposed DeiT-S-TPS with DeiT-S-2D and DeiT-S-TSM. DeiT-S-2D uses 2D DeiT-S for frame feature extraction and average pooling for temporal modeling. DeiT-S-TSM applies temporal channel shift~\cite{lin2019tsm} on both image patches and class token, and the channel shift ratio is set the same as~\cite{lin2019tsm}. The results are show in Table~\ref{table:backbone}. DeiT-S-TPS improves $2.3\%$ top-1 accuracy on Kinetics400 over DeiT-S-2D. DeiT-S-TPS also outperforms DeiT-S-TSM by $0.5\%$ thanks to the efficient spatio-temporal self-attention modeling.

	\begin{table}[h]
		\begin{center}
			\caption{More backbones experiments on Kinetics400.}
			\label{table:backbone}
			\resizebox{0.68\textwidth}{8mm}{
				\begin{tabular}{ccccccccc}
					\hline
					Model & Pretrain & Crops $\times$ Clips & FLOPs & Params & Top-1  & Top-5   \\  \hline
					
					DeiT-S-2D~\cite{touvron2021training} & IN-1K &  $ 3\times4 $ & 74G & 22M & 73.0 & 90.7 \\ 
					DeiT-S-TSM & IN-1K &  $ 3\times4 $ & 74G & 22M & 74.8 & 91.6 \\ 
					DeiT-S-TPS & IN-1K &  $ 3\times4 $ & 74G & 22M & \textbf{75.3} & \textbf{91.8} \\ \hline

				\end{tabular}
			}
		\end{center}
		
	\end{table}
	
	\begin{figure}[h]
		
		\begin{center}
			\includegraphics[width=0.7\linewidth]{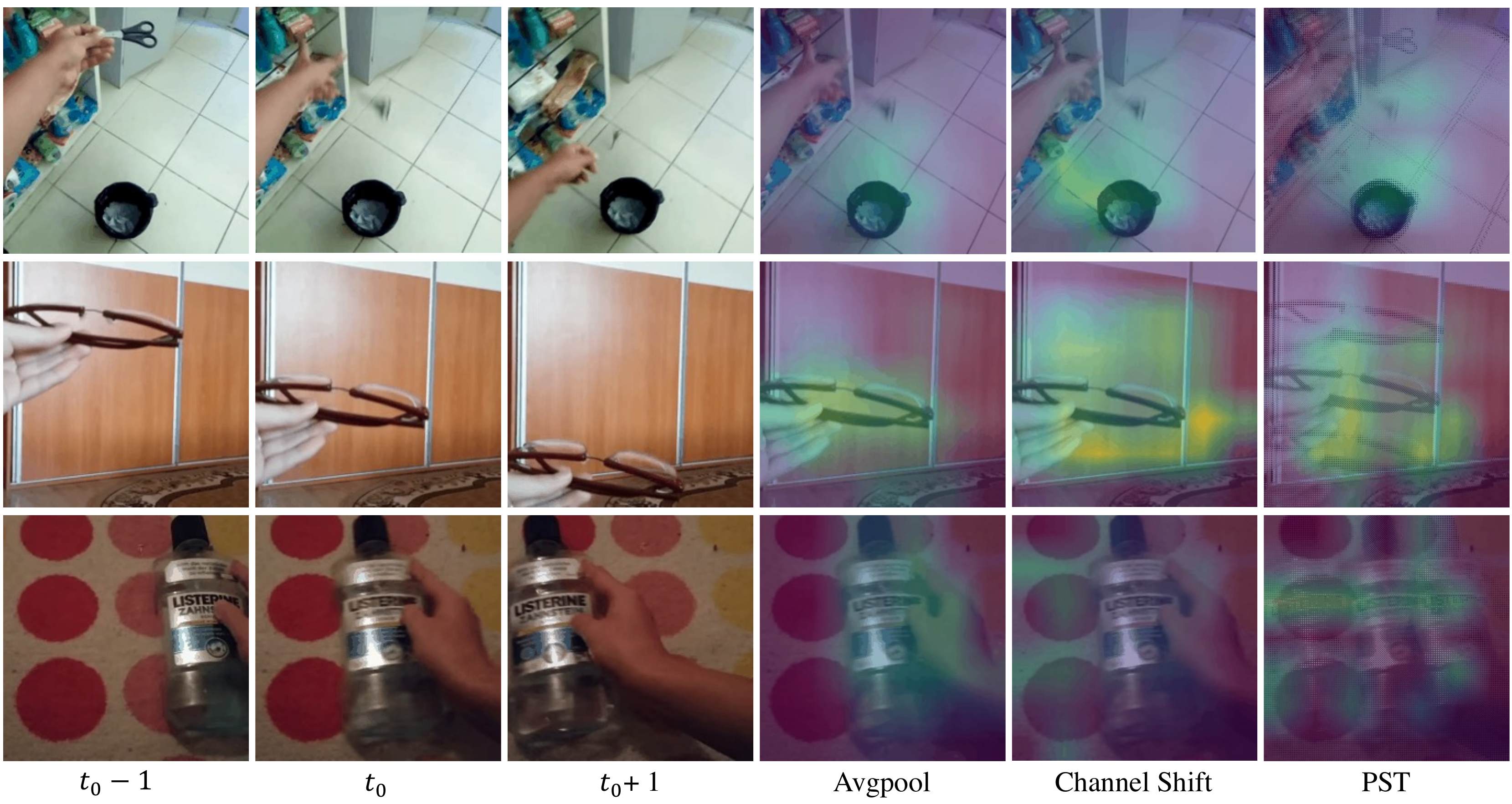}
		\end{center}
		\caption{Visualization by GradCAM on Something-something V2. Our PST learns to focus on the moving of objects. }\label{fig:tps_gradcam}
	\end{figure}
	
	\subsection{Visualization}
	We use GradCAM~\cite{gradcam} for visualization of the last stage feature maps. In Fig.~\ref{fig:tps_gradcam}, we present three examples of consecutive frames in Something-something-V2 videos in the first three columns, the learned feature activation maps by applying average pooling, channel shift and PST are presented in fourth, fifth and sixth columns, respectively. Our results show that PST learns spatiotemporal relation by associating relevant regions, which clearly indicates the motion of actions. Channel shift can also learn motion at some extent, but the activation map indicates that the receptive field is not as broad as PST. While in feature maps of average pooling, the motion can not be easily learned. 
	
	\subsection{More visualization results}
	We also present three visualization examples of PST from Kinetics400~\cite{kay2017kinetics}, Diving48~\cite{li2018resound} and Something-something V1~\cite{goyal2017something}, respectively, in Fig.~\ref{fig:tps_gradcam}. The consecutive frames of each video are shown in the first three columns, and the learned feature maps by PST are presented in the last column. It can be seen that PST can learn to associate relevant regions, and the feature maps can indicate the motion of actions in the video.
	
	\begin{figure}[h]
		
		\begin{center}
			\includegraphics[width=0.6\linewidth]{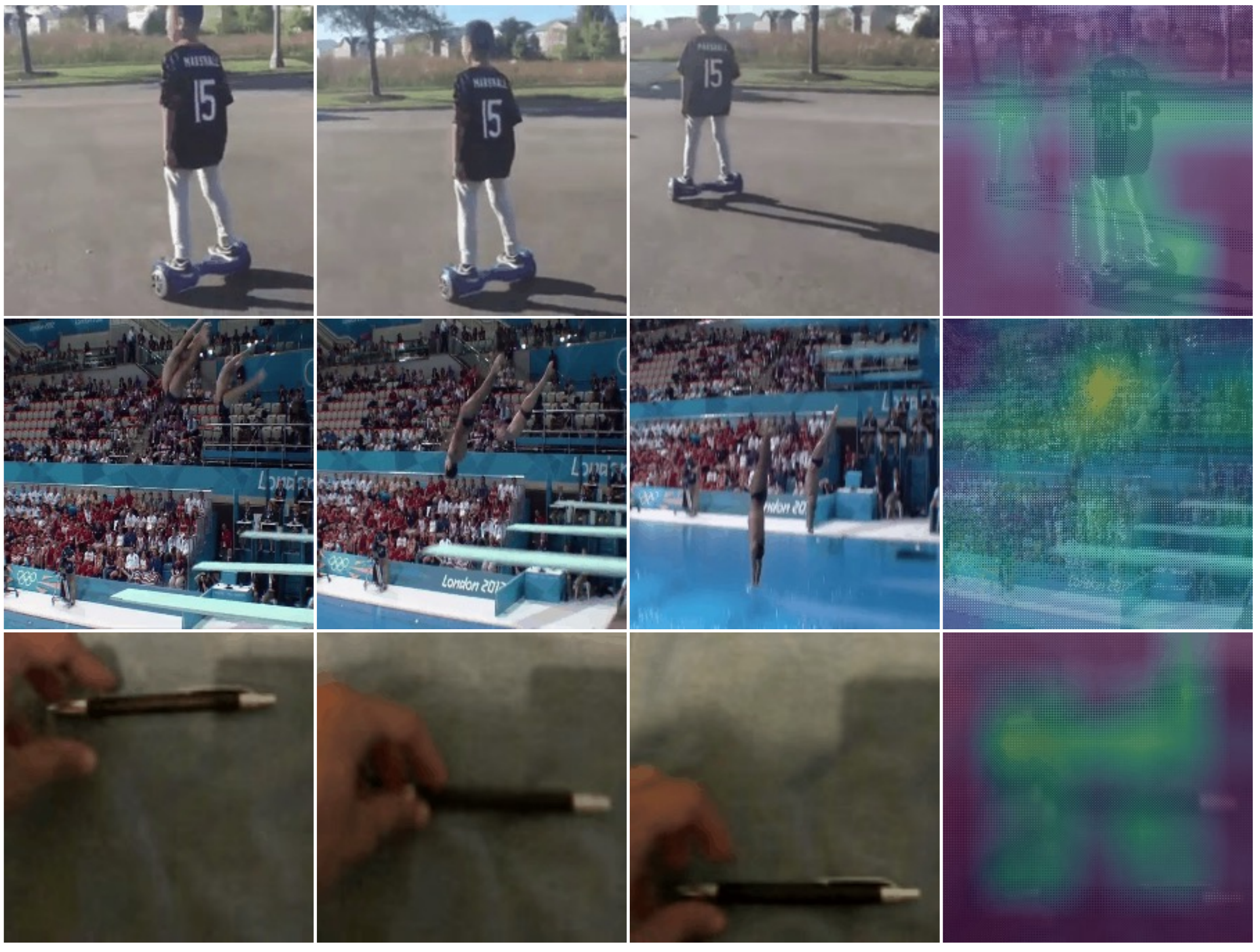}
		\end{center}
		\caption{Visualization by GradCAM on Kinetics400 (first row), Diving48 (second row) and Something-something V1 (third row. PST can learn to focus on the motion of objects. }\label{fig:tps_gradcam}
	\end{figure}

	%
	%
	\bibliographystyle{splncs04}
	\bibliography{egbib}
\end{document}